\documentclass[10pt, twocolumn, letterpaper]{article}

\usepackage{iccv}              
\usepackage{booktabs}
\usepackage{multirow}
\usepackage{xcolor}
\usepackage{colortbl}
\usepackage{makecell}

\definecolor{iccvblue}{rgb}{0.21,0.49,0.74}
\usepackage[pagebackref, breaklinks, colorlinks, allcolors=iccvblue]{hyperref}

\def\paperID{1250} 
\def\confName{ICCV}
\def\confYear{2025}

\title{OminiControl2: Efficient Conditioning for Diffusion Transformers}

\author{ Zhenxiong Tan\footnotemark[1] \quad Qiaochu Xue\thanks{Equal Contribution} \quad Xingyi Yang \quad Songhua Liu \quad
Xinchao Wang \\ National University of Singapore\\
{ \tt \small \{zhenxiong, e1352520, xyang, songhua.liu\}@u.nus.edu xinchao@nus.edu.sg }
}

\begin{document}
    \maketitle
    \begin{abstract}

    Fine-grained control of text-to-image diffusion transformer models (DiT)
    remains a critical challenge for practical deployment. While recent advances
    such as OminiControl~\cite{tan2024ominicontrol} and others have enabled a controllable
    generation of diverse control signals, these methods face significant
    computational inefficiency when handling long conditional inputs. We present
    OminiControl2, an efficient framework that achieves efficient image-conditional
    image generation. OminiControl2 introduces two key innovations: (1) a dynamic
    compression strategy that streamlines conditional inputs by preserving only
    the most semantically relevant tokens during generation, and (2) a conditional
    feature reuse mechanism that computes condition token features only once and
    reuses them across denoising steps. These architectural improvements
    preserve the original framework's parameter efficiency and multi-modal
    versatility while dramatically reducing computational costs. Our experiments
    demonstrate that OminiControl2 reduces conditional processing overhead by over
    90\% compared to its predecessor, achieving an overall 5.9× speedup in multi-conditional
    generation scenarios. This efficiency enables the practical implementation of
    complex, multi-modal control for high-quality image synthesis with DiT
    models.
\end{abstract}
    \section{Introduction}
\label{sec:intro} Diffusion Transformer (DiT) models~\cite{peebles2023scalable}
have established state-of-the-art performance in image synthesis~\cite{blackforestlabs_flux,
chen2023pixart, esser2024scaling}. The impressive performance of these models underscores
the need for control mechanisms tailored specifically to these architectures.
Although controllable generation~\cite{zhang2023adding, ye2023ip, zhang2024ssr, mou2024t2i,
li2025controlnet}
has been extensively studied for UNet-based diffusion models~\cite{rombach2021highresolution,
podell2023sdxl}, diffusion transformer models require dedicated control mechanisms
that accommodate their unique operational paradigms.

Among many attempts, OminiControl~\cite{tan2024ominicontrol} stands out as the
first versatile control framework designed for DiT. In its core, OminiControl
concatenates tokens derived from various control signals with tokens
representing noisy images, forming a single long sequence. This unified sequence
is then processed by multi-modality attention (MM-Attention)~\cite{esser2024scaling,
wei2020multi}, which enables joint modeling of multimodal inputs. Crucially, OminiControl
avoids architectural complexity by reusing the diffusion model’s pretrained VAE to
encode conditional images directly into the latent space. This approach
eliminates the need for auxiliary modules, significantly reducing the total parameter
count thus reduced the total paramter. These design choices enable OminiControl
to be easily extended to support diverse control types with minimal overhead.

Despite its advantages, this unified sequence design in OminiControl faces a significant
challenge: \emph{computational complexity}. Processing long token sequences
becomes especially problematic when the conditional sequence is extensive or
when multiple conditions (such as text, image, inpainting cues, etc.) are
applied simultaneously. Since the attention mechanism scales quadratically with sequence
length, handling multiple or lengthy conditions quickly leads to prohibitive
computational costs, limiting its practical use in complex, multi-conditional
scenarios.

To address these challenges of computational complexity, we propose \textbf{OminiControl2}\footnote{Code and more details are available at:  \url{https://github.com/Yuanshi9815/OminiControl}},
which builds on the strengths of OminiControl while specifically targeting the computational
bottlenecks especially in multi-conditioned image generation tasks. Our approach
is grounded in two core principles: reducing sequence length and minimizing redundant
computations, leading to two key innovations.

\begin{itemize}
    \item \textbf{Compact token representation.} We optimize computational
        efficiency by selectively retaining only the most informative condition
        tokens during generation. For tasks where only partial image regions require
        synthesis (e.g., localized edits), we restrict the diffusion process to
        the specific areas where new content is introduced. For full-image generation
        tasks, we condition the synthesis of high-resolution outputs on
        downsampled conditional tokens. Both strategies significantly reduce the
        number of tokens processed by the DiT during inference, streamlining
        computation without compromising quality.

    \item \textbf{Conditional feature reuse.} We compute conditional token
        features \emph{only once} during the initial inference step and reuse them
        across subsequent steps. That is to say, only the generated image token feature
        are updated across different inference steps, eliminating redundant
        computation. We analyze the feasibility of this reuse scheme,
        identifying scenarios where fixed conditional features preserve
        generation fidelity while offering computational benefits—particularly
        critical in multi-conditional generation with complex input combinations.
\end{itemize}

By integrating these advancements, OminiControl2 retains the parameter efficiency
and universal control capabilities of OminiControl while dramatically improving
computational efficiency. Our approach reduces conditional processing overhead by
over 90\% compared to the original OminiControl, enabling a 5.9× speedup in
multi-conditional scenarios while maintaining generation quality. This significant
efficiency gain enables practical implementation of complex control tasks with substantially
lower computational demands, making multi-modal control feasible even with
limited computational resources.

In summary, our contributions with OminiControl2 are as follows:
\begin{itemize}
    \item We introduce a compact image condition encoding strategy that significantly
        reduces token sequence length while retaining essential conditioning information.
        This directly addresses the substantial computational overhead caused by
        long condition sequences in multi-conditional DiT models.

    \item We implement a condition feature reuse mechanism that computes the conditional
        embedding once and caches it across denoising steps. This minimizes
        redundant computations.

    \item We show that OminiControl2 preserves the control versatility and parameter
        efficiency of its predecessor while reducing condition overhead by over 90\%.
\end{itemize}
    \section{Related Works}
\label{sec:related}

\paragraph{Diffusion models.}
Diffusion models have achieved remarkable success in text-to-image generation tasks~\cite{rombach2021highresolution,
podell2023sdxl}. To enable more fine-grained control beyond text prompts, several
works have incorporated image control signals into UNet-based diffusion models.
These approaches primarily follow two paradigms: (1) direct feature addition,
where condition features are spatially aligned and added to the hidden states of
the denoising network~\cite{zhang2023adding, mou2024t2i, qin2023unicontrol, cao2025relactrl};
and (2) cross-attention mechanisms, where separate encoders extract condition
features that are then integrated via attention operations~\cite{ye2023ip, zhang2024ssr,
he2025anystory, tian2025mige}. However, these methods are specifically designed
for UNet architectures and cannot be directly applied to Diffusion Transformer (DiT)
models due to fundamental differences in architectural design and operational
paradigms.

\paragraph{Controllable diffusion transformers.}
Recent works such as OminiControl~\cite{tan2024ominicontrol}, DSD~\cite{cai2024diffusion},
and others~\cite{mao2025ace++, she2025customvideox, cao2025relactrl} have explored
controllable generation in DiT models. These approaches elegantly leverage the
existing Multi-Modal Attention (MM-Attention) mechanism~\cite{pan2020multi}
within DiTs to incorporate image conditions without requiring complex
architectural modifications. However, they face a significant limitation: as the
number of condition tokens increases (particularly in multi-conditional
scenarios), the computational cost grows significantly due to the self-attention
operations over the entire token sequence, making them inefficient for practical
applications with multiple or high-resolution condition inputs.

\paragraph{Acceleration techniques for diffusion transformers.}
The iterative nature of diffusion models incurs substantial computational costs.
Recent works have focused on enhancing DiT's efficiency through various techniques:
\textit{Model Pruning} approaches dynamically skip or remove components of the model.
DiP-GO~\cite{zhu2024dipgodiffusionprunerfewstep} predicts whether to skip
computational blocks during inference, while TinyFusion~\cite{fang2024tinyfusiondiffusiontransformerslearned}
removes redundant layers from diffusion transformers. \textit{Computation
Caching} stores intermediate results to reduce redundancy. FORA~\cite{selvaraju_fora_2024}
reuses intermediate outputs from attention and MLP layers across denoising steps.
TokenCache~\cite{lou_token_2024} and related methods~\cite{chen2024deltadittrainingfreeaccelerationmethod,
zou_token_accelerating_2025,ma_learning--cache_2024, zou2024accelerating,
shen2024lazydit} achieve efficiency by strategically caching less important tokens.
These works provide valuable insights for our work, which focuses specifically on
reducing computational overhead in multi-conditional DiT models while maintaining
control versatility and generation quality.
    \section{Methods}
\label{sec:methods}
\subsection{Preliminaries}

\paragraph{Diffusion transformer.}
The text-guided Diffusion Transformer (DiT) models~\cite{peebles2023scalable,blackforestlabs_flux,
esser2024scaling, chen2023pixart}
generates high-quality images through an iterative denoising process. In the model
of FLUX~\cite{blackforestlabs_flux}, at each denoising step, transformer blocks process
a token sequence
\begin{equation}
    \mathbf{S}= [\mathbf{X}; \mathbf{C}_{T}]
\end{equation}
This sequence consists of noisy tokens $\mathbf{X}\in \mathbb{R}^{* \times d}$ and
text condition tokens $\mathbf{C}_{T}\in \mathbb{R}^{* \times d}$, where $d$ is
the embedding dimension, and $*$ denotes the number of tokens.

\paragraph{Unified token sequence.}
The previous work~\cite{tan2024ominicontrol} extends DiT's input sequence by
introducing image condition tokens. For an input condition image, it is first
encoded into latent tokens $\mathbf{C}_{I}\in \mathbb{R}^{* \times d}$ using the
VAE encoder. The extended token sequence becomes:
\begin{align}
    \mathbf{S} & = [\mathbf{X}; \mathbf{C}_{T}; \mathbf{C}_{I}]
\end{align}
This unified token framework enables flexible interactions between image and
condition tokens through transformer layers, supporting various image generation
tasks.

\paragraph{Position index in controllable DiT.}
For each token in the sequence, there is a corresponding position index ${P}\in \mathbb{R}
^{2}$, which is involved in attention computation to capture spatial
dependencies~\cite{blackforestlabs_flux, su2024roformer}. Specifically, the
position index for noisy tokens is determined by the spatial position $(i, j)$
in the 2D image grid, while the position index for text condition tokens is set to
a fixed value $(0, 0)$.

For image condition tokens, the position index varies depending on the condition
task type. For spatially aligned imgae condition tasks (e.g., edge-guided, depth-guided
generation), position indices share the same spatial positions $(i, j)$ as the corresponding
noisy tokens. For spatially non-aligned tasks (e.g., subject-driven generation),
position indices are shifted by a fixed offset $(\Delta i, \Delta j)$. This
adaptive position indexing strategy~\cite{tan2024ominicontrol} enables the model
to effectively handle the spatial relationships between different condition
types.

\paragraph{Extension to multi-image condition tasks.}
To extend this controllable DiT formulation for multi-image condition tasks, we
can simply concatenate multiple image condition tokens
$\mathbf{C}_{I}^{(k)}\in \mathbb{R}^{* \times d}$ as follows:
\begin{align}
    \mathbf{C}_{I} & = [\mathbf{C}_{I}^{(1)}; \mathbf{C}_{I}^{(2)}; ...; \mathbf{C}_{I}^{(K)}]
\end{align}
where $K$ is the number of condition images. While this approach is
straightforward and easily implemented, it leads to a large number of condition tokens
and increase the computational cost of the model.

\subsection{Compact token representation}
To address the issue of increasing computational cost in multi-image condition
tasks, we first try to reduce the number of condition tokens. Specifically, we propose
a compact encoding method based on compression and pruning.

\paragraph{Compression with position correcting.}

\begin{figure}[t]
    \centering
    \includegraphics[width=\linewidth]{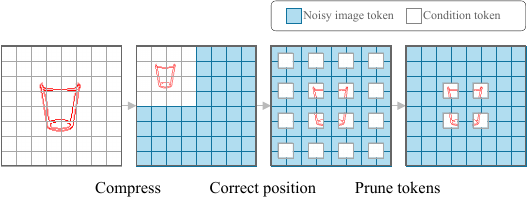}
    \caption{Illustration of the compression and position correcting of compact
    token representation for condition images.}
    \vspace{-1em}
    \label{fig:radar}
\end{figure}

We implement a spatial compression strategy for condition images prior to VAE
encoding, reducing spatial dimensions by a factor of $a$ in each direction. This
transformation achieves a $a^{2}:1$ reduction in token count—converting an
$n \times n$ token field into a more efficient $\frac{n}{a}\times \frac{n}{a}$
representation (see Figure~\ref{fig:radar}).

However, naïve compression creates spatial misalignment between the compressed condition
space and the target generation space. For spatially-aligned tasks, this misalignment
leads to structural inconsistencies and degraded control precision, as
compressed condition tokens no longer maintain direct correspondence with noisy
image tokens.

To preserve spatial coherence, we introduce an position correcting function that
establishes correspondence between the compressed condition image tokens and their
target regions in the generated image:
\begin{equation}
    P_{C_I}(i,j) \mapsto P_{X}(a \cdot i, a \cdot j)
\end{equation}

This mapping is crucial and ensures each compressed token provides guidance to its
appropriate region in the generated image, maintaining conditioning fidelity
while substantially reducing computational requirements. Without this correction,
the model would struggle to learn the spatial relationships between condition
and generated tokens, leading to poor quality results (Section
\ref{sec:exp:empirical}).

\paragraph{Token pruning.}
We also implement a token pruning strategy to eliminate non-informative tokens
from condition images. This approach identifies and removes tokens that contribute
minimally to the conditioning signal, particularly effective for sparse
conditional inputs.

Token relevance is determined by condition-specific criteria. For example, in
edge-guided generation, tokens representing regions without edges (values near
zero) are pruned as they provide little guidance information. The position
information of retained tokens remains unchanged, preserving spatial correspondence
after pruning.

\begin{figure}[t]
    \centering
    \includegraphics[width=\linewidth]{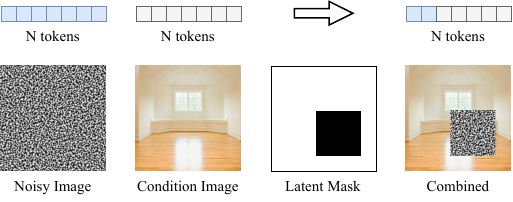}
    \caption{Illustration of token integration processing for inpainting. By combining
    noisy and condition tokens based on the mask, we reduce token count from 2N
    to N.}
    \vspace{-1em}
    \label{fig:fill}
\end{figure}



\paragraph{Token integration.}
\label{sec:token_integration} For inpainting tasks, we propose a novel token
integration methodology that incorporates condition tokens as intrinsic
components of the output token representation (Figure~\ref{fig:fill}). Instead of
maintaining discrete token sequences for condition images, we strategically
integrate condition tokens $\mathbf{C}_{I}^{\mathbf{M}=0}$ (corresponding to unmasked
regions) as invariant elements within the output token space. This approach preserves
the computational processing of these tokens through transformer layers at each denoising
step, while conceptually designating them as output tokens exempt from
modification. By selectively constraining the denoising process to operate exclusively
on tokens within the masked region $\mathbf{X}^{\mathbf{M}=1}$, while
simultaneously preserving unmasked areas as static contextual information, we maintain
the critical spatial relationships necessary for high-fidelity synthesis. The
resultant output constitutes a coherent integration of newly synthesized content
within masked regions alongside preserved original content in unmasked areas. This
methodology ensures complete fidelity preservation in the VAE's latent representation
while optimizing computational efficiency in the processing pipeline.

\subsection{Feature reuse in DiT}
Another key factor contributing to the computational complexity of DiT models is
the repetitive processing of condition tokens across multiple denoising steps. In
the standard diffusion process, at each denoising timestep $t$, the DiT model processes
a token sequence that includes both the current noisy latent tokens and various condition
tokens~\cite{peebles2023scalable, blackforestlabs_flux}. However, we observe a fundamental
asymmetry in how these elements evolve during inference: while the noisy latent
representation changes progressively as noise is removed across timesteps, the image
condition inputs remain unchanged throughout the entire sampling process~\cite{tan2024ominicontrol}.
This observation suggests an opportunity for significant computational
optimization—what if we could compute the intermediate features for condition
tokens just once and reuse them across all denoising steps?

\begin{figure}[t]
    \centering
    \includegraphics[width=\linewidth]{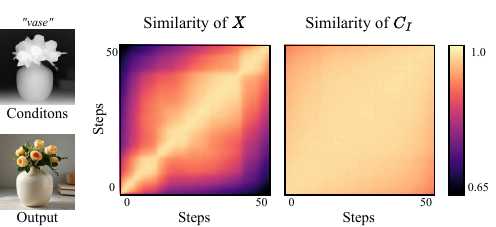}
    \caption{Feature similarity across denoising steps. While noisy image tokens
    $X$ change significantly between steps (left), condition tokens $C_{I}$
    maintain high similarity throughout the denoising process (right).}
    \label{fig:feature_similarity}
\end{figure}

To validate this, we conduct an empirical analysis by measuring the cosine
similarity of token features across different timesteps. As shown in Figure~\ref{fig:feature_similarity},
while noisy image tokens $\mathbf{X}$ exhibit substantial feature variation
between timesteps, condition tokens $\mathbf{C}_{I}$ maintain remarkably high similarity
throughout the diffusion process. This high similarity indicates significant
computational redundancy when recalculating condition features at each step.

\paragraph{Naïve cache strategy.}
\label{sec:feature_caching}

Inspired by caching techniques in large language models (LLMs)~\cite{pope2022efficiently}
and DiT~\cite{chen2024deltadittrainingfreeaccelerationmethod, selvaraju_fora_2024,
sun_unicp_2025, ma2025learning}, we initially attempted an approach where condition
features are computed only once during the first inference step and then reused across
subsequent steps. Since the interaction between condition tokens $\mathbf{C}$
and noisy tokens $\mathbf{X}$ occurs exclusively within the MM-attention modules~\cite{pan2020multi,
tan2024ominicontrol}, we specifically reuse the key-value ($KV$) projections of condition
tokens (see Figure~\ref{fig:reuse}(a, b)).

\begin{figure}[t]
    \centering
    \vspace{1em}
    \includegraphics[width=\linewidth]{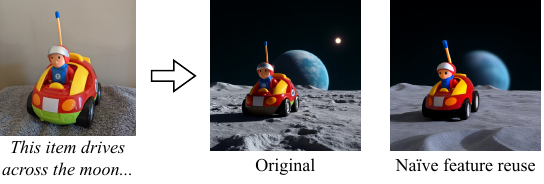}
    \caption{Visual comparison of the original inference pipeline (left) and
    naive feature reuse strategy (right). }
    \label{fig:reuse}
\end{figure}

However, despite the high similarity observed in Figure~\ref{fig:feature_similarity},
this approach produced unsatisfactory results as shown in Figure~\ref{fig:reuse}.
This performance gap reveals that while condition features remain largely consistent,
they still undergo subtle dynamic changes during the denoising process. In our naive
implementation, condition features remain entirely static, creating a training-inference
discrepancy that degrades output quality.

\paragraph{Asymmetric attention masking.}
To resolve this discrepancy, we need to ensure that condition token features remain
consistent across denoising steps. Unlike LLMs, which can effectively implement KV-caching
due to their causal attention mechanism (where subsequent tokens cannot
influence preceding ones), DiT employs full multi-modal attention where changes
in noisy tokens $\mathbf{X}$ affect condition tokens $\mathbf{C}$ and vice versa~\cite{tan2024ominicontrol,
esser2024scaling}.

\begin{figure}[t]
    \centering
    \includegraphics[width=\linewidth]{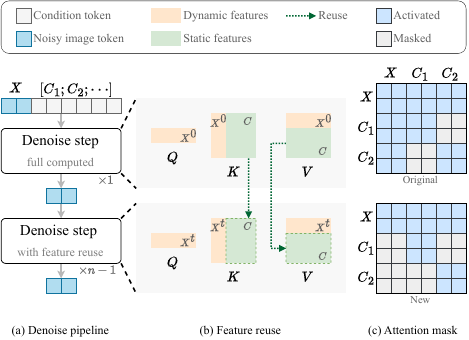}
    \caption{Illustration of our feature reuse strategy. (a) Overview of the denoising
    pipeline, with full computation performed only at the first step. (b) Detailed
    view of the feature reuse mechanism in Attention, where condition token
    features($K$$V$) computed in the first step are reused in subsequent steps. (c)
    Asymmetric attention mask that prevents condition tokens from attending to noisy
    image tokens, enabling consistent feature reuse.}
    \label{fig:reuse}
\end{figure}

We address this by introducing an asymmetric attention mechanism for DiT, as
illustrated in Figure~\ref{fig:reuse}(c). In this modified approach, noisy image
tokens $\mathbf{X}$ retain their original attention pattern, attending to both
image and condition tokens $[\mathbf{X}; \mathbf{C}]$. However, we apply
attention masking to prevent condition tokens $\mathbf{C}$ from attending to
noisy image tokens $\mathbf{X}$. This asymmetric setup, incorporated during
model training, ensures that while image tokens benefit from conditioning information,
the condition token features remain independent of the evolving noisy image
representation. This alignment between training and inference enables effective feature
reuse, allowing us to compute condition features just once while maintaining
generation quality.

\subsection{Computational complexity analysis}
\label{subsec:complexity}

To quantify the efficiency gains of our approach, we analyze the computational
complexity of controllable DiT models and identify key optimization
opportunities. For each image generation process with the original method, the
time complexity can be formulated as:
\[
    O(n \cdot (d^{2}N + N^{2}d))
\]
where $n$ is the number of denoising steps, $N = |X| + |C|$ is the total token
count (image and condition tokens combined), and $d$ is the feature dimension.
We can simplify this expression to:
\[
    O(n \cdot (c_{1}N + c_{2}N^{2}))
\]
where $c_{1}N$ represents token-independent operations (projections, MLPs,
normalization) and $c_{2}N^{2}$ represents the attention computation between tokens.

\begin{figure}[t]
    \centering
    \includegraphics[width=1.0\linewidth]{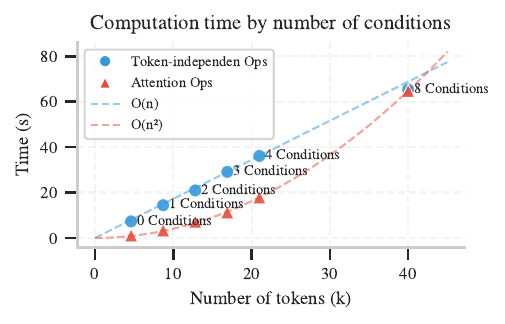}
    \vspace{-2em}
    \caption{Computational time breakdown analysis. For typical scenarios (1-4
    conditions), token-independent operations (blue circles) dominate the
    computational cost, while attention operations (red triangles) become significant
    only with higher condition counts (8). X-axis shows token count ($|X|+|C|$)
    in thousands. }
    \label{fig:complexity}
\end{figure}

Our empirical measurements in Figure~\ref{fig:complexity} reveal that for
practical use cases with 1-4 condition sources, the computational overhead is dominated
by token-independent operations rather than the quadratic attention computations.
Under this, the computational overhead introduced by condition tokens can be approximated
as $O(n \cdot |C|)$, where $|C|$ is the number of condition tokens. Our proposed
methods address this overhead from two complementary directions:

\textbf{Compact token representation} reduces condition token count from $|C|$
to $r \cdot |C|$ (where $r < 1$), providing a speedup factor for condition-related
overhead of:
\[
    \boldsymbol{\alpha}_{\text{compact}}= \frac{1}{r}
\]

\textbf{Conditional token reuse} computes condition token features once instead
of $n$ times, yielding a speedup factor for condition processing of:
\[
    \boldsymbol{\alpha}_{\text{reuse}}= n
\]

When combined, these approaches deliver a cumulative theoretical speedup for
condition-related computational overhead of:
\[
    \boldsymbol{\alpha}_{\text{total}}= \frac{n}{r}
\]

For a common configuration with $n=28$ and $r=0.25$, this yields a theoretical speedup
of $\boldsymbol{\alpha}_{\text{total}}=112$ for condition token processing. For
a step-distilled model with $n=4$ and $r=0.25$, the speedup factor is $\boldsymbol
{\alpha}_{\text{total}}=16$.
    \section{Experiments}
\label{sec:exp}

\subsection{Setup}
\label{subsec:setup}

\paragraph{Base model and training.}
We adopt FLUX.1~\cite{blackforestlabs_flux} as our base model. Following
OminiControl~\cite{tan2024ominicontrol}, we fine-tune using LoRA~\cite{devalal2018lora}
with identical rank and parameter settings. Training employs the Prodigy
optimizer~\cite{mishchenko2024prodigy} for 25k iterations per task, with batch size
1 and gradient accumulation over 4 steps. All experiments are conducted on a
single NVIDIA H100 GPU. The training dataset is Text-to-Image-2M~\cite{jackyhate2024t2i}
following the same setup as OminiControl~\cite{tan2024ominicontrol}.

\newpage
\textbf{Tasks.} We evaluate our methods on four conditional generation tasks: Canny-to-image,
depth-to-image, mask inpainting, and image deblurring. Additionally, we assess
performance in a multi-condition setting where all four condition types are
simultaneously applied.

\textbf{Implementation Details.} For our compact token representation, task-specific
strategies are implemented: mask inpainting adopts the approach of token integration
(Section~\ref{sec:token_integration}), while other tasks use spatial compression
with $r=0.25$, reducing token dimensions by 50\% in each direction. Additionally,
the non-informative tokens in canny edge maps are pruned.

\textbf{Baselines.} We first compare against the original OminiControl~\cite{tan2024ominicontrol}
as our primary baseline. We also compare against the ControlNet~\cite{zhang2023adding}
which is implemented on StableDiffusion 1.5~\cite{rombach2021highresolution} and
ControlNetPro~\cite{flux1controlnet2024} which is implemented on FLUX.1. For
OminiControl~\cite{tan2024ominicontrol}, we trained the model with the same training
data and settings as our method. Additionally, we evaluate against two efficiency-focused
methods for DiT models: token-merging techniques~\cite{bolya2022token, bolya2023token}
\footnote{ Due to constraints imposed by positional embeddings in FLUX~\cite{blackforestlabs_flux},
token-merging approaches can only be applied to MLP and feedforward components
while preserving the original attention mechanisms. } that reduce computation by
strategically combining similar tokens, and naive feature caching that introduced
in Section~\ref{sec:feature_caching}.

\textbf{Evaluation.} Efficiency metrics include inference latency and condition
overhead, measured on an single NVIDIA RTX 6000 Ada GPU. Generation quality is
evaluated using FID~\cite{heusel2017gans}, CLIP~\cite{hessel2021clipscore}, NIQE~\cite{mittal2012making},
and MUSIQ~\cite{ke2021musiq}. We use total 5,000 images from COCO 2017
validation set, and resize them to 512$\times$512, then generate images with
task-specific conditions and associated captions as prompts with a fixed seed of
42.

\subsection{Main results}

\begin{figure}[!t]
    \centering
    \includegraphics[width=\linewidth]{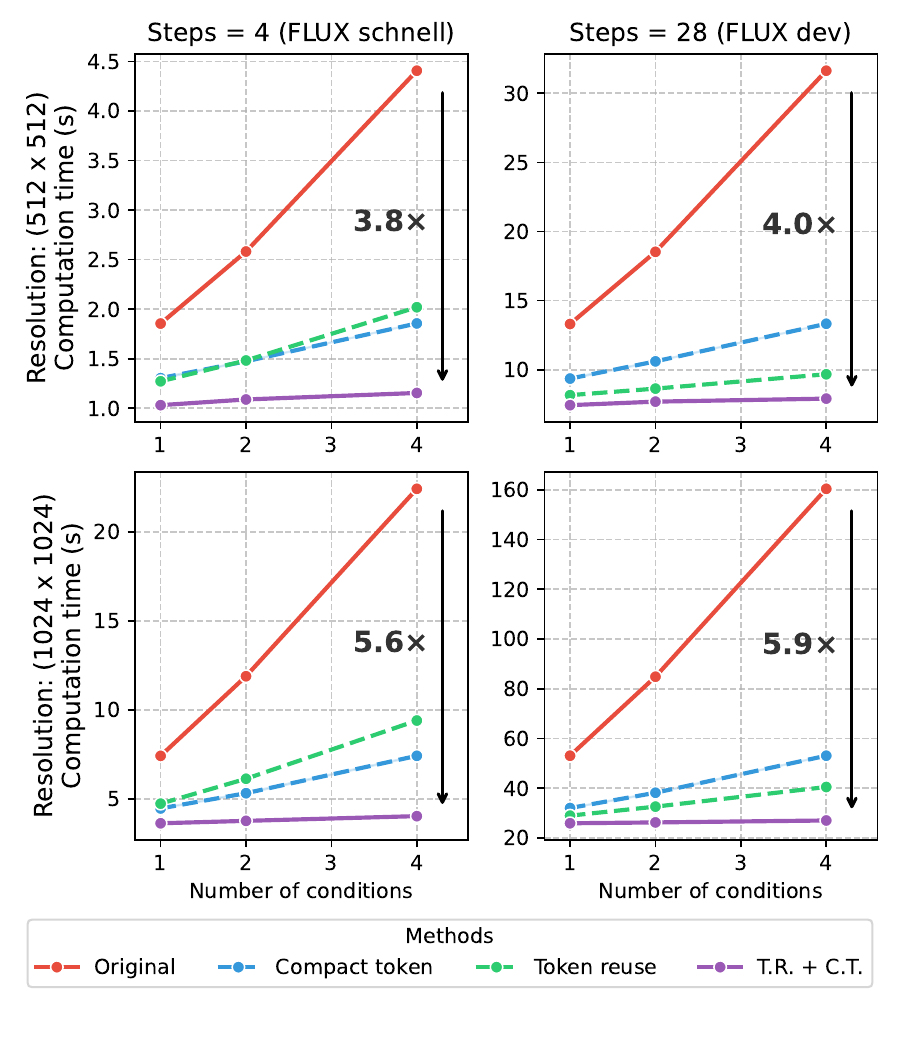}
    \caption{Computation time comparison across different resolution and
    sampling step configurations. Our combined approach (T.R. + C.T.) achieves 3.8-5.9×
    speedup over the original method, with greater gains at higher resolutions and
    more conditions. Both individual optimizations show significant improvements,
    but their combination delivers maximum efficiency. (Compact token representation
    reduces token count by approximately 75\%.)}
    \label{fig:efficiency}
\end{figure}

\paragraph{Efficiency.}
Figure~\ref{fig:efficiency} illustrates computation times across different configurations.
The experimental results show that our combined approach offers speedups ranging
from 3.8$\times$ to 5.9$\times$ compared to the original OminiControl
implementation. We observe that each optimization strategy contributes differently
depending on the configuration: compact token representation tends to be more
effective in low-step settings (FLUX schnell with 4 denoising steps), while token
reuse shows greater efficiency gains with more steps (FLUX dev with 28 denoising
steps). Notably, the measurements indicate that our combined method maintains relatively
consistent computation time as the number of conditions increases, whereas the
baseline method exhibits a more linear growth pattern. This efficiency characteristic
becomes particularly valuable at higher resolutions (1024×1024), potentially
making multi-conditional generation more practical in computationally intensive
scenarios.

\begin{figure*}[!t]
    \centering
    \includegraphics[width=\linewidth]{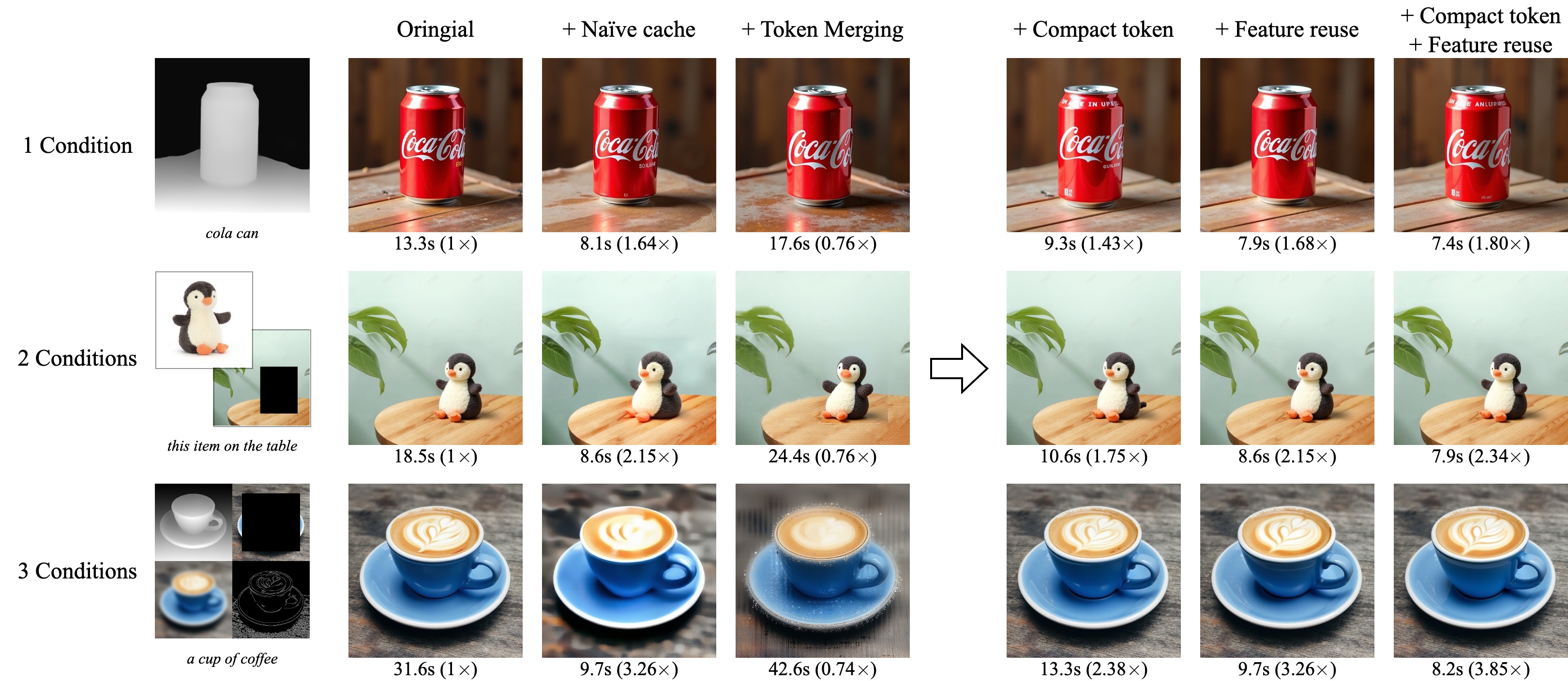}
    \caption{Performance comparison of different methods with varying condition
    counts for 512×512 image generation. Our approach maintains consistent quality
    with increased conditions while achieving the fastest inference speeds (up
    to 3.85× speedup), whereas naive caching and token merging show deteriorating
    performance as condition count increases. }
    \label{fig:all_comp}
\end{figure*}

\paragraph{Qualitative comparison.}
Figure~\ref{fig:all_comp} compares generation quality and inference time across
different optimization approaches with varying numbers of conditions. Both our compact
token representation and feature reuse methods produce high-quality images while
significantly reducing generation time. Notably, even as the number of conditions
increases from 1 to 4, our methods maintain consistent image quality while
achieving substantial speedups (up to 3.85× with the combined approach). In
contrast, baseline methods such as naïve cache and token merging show clear degradation
in image quality when handling multiple conditions, particularly with 4
conditions where visual artifacts become apparent. We observe that token-merge approaches
face challenges in the FLUX architecture, as they require additional
computational overhead for token grouping operations, and their effectiveness is
constrained by the position embedding mechanism. This may explain why token-merge
methods, despite compromising on image quality, do not deliver the performance
improvements that might be expected in this particular context.

\paragraph{Quantitative evaluation.}

\begin{table*}
    [!t]
    \centering
    \small
    \begin{tabular}{l|l|ccccc|cc}
        \toprule \multirow{2}{*}{Task}                          & \multirow{2}{*}{Method} & \multirow{2}{*}{FID $\downarrow$}      & \multirow{2}{*}{CLIP-Text $\uparrow$} & \multirow{2}{*}{CLIP-Image $\uparrow$} & \multirow{2}{*}{NIQE $\downarrow$}    & \multirow{2}{*}{MUSIQ $\uparrow$}       & \multirow{2}{*}{Latency$\downarrow$} & \multirow{2}{*}{\makecell{Condition\\overhead (s)}$\downarrow$} \\
                                                                &                         &                                        &                                       &                                        &                                       &                                         &                                      &                                                                 \\
        \midrule \multirow{5}{*}{Depth}                         & OminiControl            & 26.879\cellcolor{lightgray!50}         & 0.308\cellcolor{lightgray!50}         & 0.745\cellcolor{lightgray!50}          & 4.247\cellcolor{lightgray!50}         & 73.498\cellcolor{lightgray!50}          & 13.30\cellcolor{lightgray!50}        & 6.05 / 9.9\%\cellcolor{lightgray!50}                            \\
                                                                & Naïve cache             & 28.885 \cellcolor{iccvblue!15}         & 0.306 \cellcolor{iccvblue!15}         & \textbf{0.743} \cellcolor{iccvblue!15} & 4.597 \cellcolor{iccvblue!15}         & 73.349 \cellcolor{iccvblue!15}          & 8.16 \cellcolor{iccvblue!15}         & 0.91 / 15.0\% \cellcolor{iccvblue!15}                           \\
                                                                & Ours                    & \textbf{28.719}\cellcolor{iccvblue!15} & \textbf{0.308}\cellcolor{iccvblue!15} & 0.731\cellcolor{iccvblue!15}           & \textbf{4.296}\cellcolor{iccvblue!15} & \textbf{73.669}\cellcolor{iccvblue!15}  & 7.85\cellcolor{iccvblue!15}          & 0.60 / 9.9\% \cellcolor{iccvblue!15}                            \\
                                                                & Ours / F.R. only        & 28.719                                 & 0.308                                 & 0.731                                  & 4.296                                 & 73.669                                  & 8.18                                 & 0.93 / 15.3\%                                                   \\
                                                                & Ours / C.T.R only       & 27.730                                 & 0.308                                 & 0.741                                  & 4.244                                 & 72.470                                  & 9.38                                 & 2.13 / 35.2\%                                                   \\
        \midrule \multirow{4}{*}{Inpainting}                    & OminiControl            & 12.078\cellcolor{lightgray!50}         & 0.304\cellcolor{lightgray!50}         & 0.884\cellcolor{lightgray!50}          & 3.781\cellcolor{lightgray!50}         & 71.889\cellcolor{lightgray!50}          & 13.30\cellcolor{lightgray!50}        & 6.05 / 100\%\cellcolor{lightgray!50}                            \\
                                                                & Naïve cache             & 16.727 \cellcolor{iccvblue!15}         & 0.296 \cellcolor{iccvblue!15}         & 0.854 \cellcolor{iccvblue!15}          & 4.204 \cellcolor{iccvblue!15}         & 64.955 \cellcolor{iccvblue!15}          & 8.16 \cellcolor{iccvblue!15}         & 0.91 / 15.0\% \cellcolor{iccvblue!15}                           \\
                                                                & Ours                    & \textbf{11.712}\cellcolor{iccvblue!15} & \textbf{0.307}\cellcolor{iccvblue!15} & \textbf{0.891}\cellcolor{iccvblue!15}  & \textbf{3.878}\cellcolor{iccvblue!15} & \textbf{71.045}\cellcolor{iccvblue!15}  & 7.85\cellcolor{iccvblue!15}          & 0.60 / 9.9\%\cellcolor{iccvblue!15}                             \\
                                                                & Ours / F.R. only        & 12.672                                 & 0.305                                 & 0.876                                  & 3.741                                 & 72.732                                  & 8.18                                 & 0.93 / 15.3\%                                                   \\
        \midrule \multirow{5}{*}{Deblur}                        & OminiControl            & 16.381\cellcolor{lightgray!50}         & 0.303\cellcolor{lightgray!50}         & 0.851\cellcolor{lightgray!50}          & 4.060\cellcolor{lightgray!50}         & 71.796\cellcolor{lightgray!50}          & 13.30\cellcolor{lightgray!50}        & 6.05 / 100\%\cellcolor{lightgray!50}                            \\
                                                                & Naïve cache             & 45.039 \cellcolor{iccvblue!15}         & 0.285 \cellcolor{iccvblue!15}         & 0.773 \cellcolor{iccvblue!15}          & 5.572 \cellcolor{iccvblue!15}         & 55.700 \cellcolor{iccvblue!15}          & 8.16 \cellcolor{iccvblue!15}         & 0.91 / 15.0\% \cellcolor{iccvblue!15}                           \\
                                                                & Ours                    & \textbf{20.775}\cellcolor{iccvblue!15} & \textbf{0.302}\cellcolor{iccvblue!15} & \textbf{0.809}\cellcolor{iccvblue!15}  & \textbf{4.856}\cellcolor{iccvblue!15} & \textbf{69.474}\cellcolor{iccvblue!15}  & 7.85\cellcolor{iccvblue!15}          & 0.60 / 9.9\%\cellcolor{iccvblue!15}                             \\
                                                                & Ours / F.R. only        & 18.638                                 & 0.303                                 & 0.836                                  & 4.014                                 & 71.863                                  & 8.18                                 & 0.93 / 15.3\%                                                   \\
                                                                & Ours / C.T.R only       & 19.664                                 & 0.303                                 & 0.831                                  & 4.204                                 & 72.260                                  & 9.38                                 & 2.13 / 35.2\%                                                   \\
        \midrule\multirow{5}{*}{Canny}                          & OminiControl            & \cellcolor{lightgray!50} 22.836        & 0.307 \cellcolor{lightgray!50}        & 0.780 \cellcolor{lightgray!50}         & 4.002 \cellcolor{lightgray!50}        & 74.570 \cellcolor{lightgray!50}         & 13.30\cellcolor{lightgray!50}        & 6.05 / 9.9\%\cellcolor{lightgray!50}                            \\
                                                                & Naïve cache             & 31.497 \cellcolor{iccvblue!15}         & 0.302 \cellcolor{iccvblue!15}         & \textbf{0.748} \cellcolor{iccvblue!15} & 5.198 \cellcolor{iccvblue!15}         & \textbf{75.128} \cellcolor{iccvblue!15} & 8.16 \cellcolor{iccvblue!15}         & 0.91 / 15.0\% \cellcolor{iccvblue!15}                           \\
                                                                & Ours                    & \textbf{29.766}\cellcolor{iccvblue!15} & \textbf{0.308}\cellcolor{iccvblue!15} & 0.724\cellcolor{iccvblue!15}           & \textbf{4.069}\cellcolor{iccvblue!15} & 73.316\cellcolor{iccvblue!15}           & 7.85\cellcolor{iccvblue!15}          & 0.60 / 9.9\%\cellcolor{iccvblue!15}                             \\
                                                                & Ours / F.R. only        & 22.274                                 & 0.306                                 & 0.758                                  & 4.150                                 & 74.483                                  & 8.18                                 & 0.93 / 15.3\%                                                   \\
                                                                & Ours / C.T.R only       & 26.654                                 & 0.309                                 & 0.742                                  & 3.979                                 & 74.094                                  & 9.38                                 & 2.13 / 35.2\%                                                   \\
        \midrule \multirow{5}{*}{\makecell[l]{Multi-condition}} & OminiControl            & 7.296\cellcolor{lightgray!50}          & 0.301\cellcolor{lightgray!50}         & 0.924\cellcolor{lightgray!50}          & 3.878\cellcolor{lightgray!50}         & 70.961\cellcolor{lightgray!50}          & 31.63\cellcolor{lightgray!50}        & 24.38 / 100\%\cellcolor{lightgray!50}                           \\
                                                                & Naïve cache             & 19.965\cellcolor{iccvblue!15}          & 0.292 \cellcolor{iccvblue!15}         & 0.876 \cellcolor{iccvblue!15}          & 4.236 \cellcolor{iccvblue!15}         & 63.984 \cellcolor{iccvblue!15}          & 9.66 \cellcolor{iccvblue!15}         & 2.41 / 9.8\% \cellcolor{iccvblue!15}                            \\
                                                                & Ours                    & \textbf{10.718}\cellcolor{iccvblue!15} & \textbf{0.303}\cellcolor{iccvblue!15} & \textbf{0.903}\cellcolor{iccvblue!15}  & \textbf{3.872}\cellcolor{iccvblue!15} & \textbf{71.273}\cellcolor{iccvblue!15}  & 8.18\cellcolor{iccvblue!15}          & 0.93 / 3.8\% \cellcolor{iccvblue!15}                            \\
                                                                & Ours / F.R. only        & 8.492                                  & 0.302                                 & 0.920                                  & 3.846                                 & 70.099                                  & 9.78                                 & 2.53 / 10.4\%                                                   \\
                                                                & Ours / C.T.R only       & 8.409                                  & 0.302                                 & 0.919                                  & 3.846                                 & 71.024                                  & 13.32                                & 6.07 / 24.9\%                                                   \\
        \bottomrule
    \end{tabular}
    \caption{Comparison of efficiency and generation quality across various
    conditioning tasks. \textbf{Bold} indicates best results between compared
    methods (blue-highlighted rows). White rows show ablation studies: F.R. = Feature
    Reuse only, C.T.R. = Compact Token Representation only. The rightmost column
    shows condition overhead in seconds / percentage relative to original method
    (OminiControl).}
    \label{tab:quantitative}
\end{table*}

\begin{table*}
    [!t]
    \centering
    \small
    \begin{tabular}{l|l|ccccc|cc}
        \toprule \multirow{2}{*}{Task} & \multirow{2}{*}{Method} & \multirow{2}{*}{FID $\downarrow$} & \multirow{2}{*}{CLIP-Text $\uparrow$} & \multirow{2}{*}{CLIP-Image $\uparrow$} & \multirow{2}{*}{NIQE $\downarrow$}    & \multirow{2}{*}{MUSIQ $\uparrow$}      & \multirow{2}{*}{\makecell{Parameters\\overhead}$\downarrow$} & \multirow{2}{*}{\makecell{Condition\\overhead (s)}$\downarrow$} \\
                                       &                         &                                   &                                       &                                        &                                       &                                        &                                                              &                                                                 \\
        \midrule\multirow{3}{*}{Depth} & SD 1.5 + ControlNet     & \textbf{23.029}                   & 0.308                                 & 0.7264                                 & 4.5802                                & 70.73                                  & 361M                                                         & 0.52                                                            \\
                                       & FLUX + ControlNet       & 62.203                            & 0.212                                 & 0.5477                                 & 4.0151                                & 66.849                                 & 3,300M                                                       & 1.16                                                            \\
                                       & Ours                    & 28.719\cellcolor{iccvblue!15}     & \textbf{0.308}\cellcolor{iccvblue!15} & \textbf{0.731}\cellcolor{iccvblue!15}  & 4.296\cellcolor{iccvblue!15}          & \textbf{73.669}\cellcolor{iccvblue!15} & 14.5M\cellcolor{iccvblue!15}                                 & 0.60 \cellcolor{iccvblue!15}                                    \\
        \midrule\multirow{3}{*}{Canny} & SD 1.5 + ControlNet     & \textbf{18.74 }                   & 0.305                                 & \textbf{0.752}                         & 4.751                                 & 67.899                                 & 361M                                                         & 0.52                                                            \\
                                       & FLUX + ControlNet       & 98.689                            & 0.192                                 & 0.537                                  & 4.636                                 & 56.907                                 & 3,300M                                                       & 1.16                                                            \\
                                       & Ours                    & 29.766\cellcolor{iccvblue!15}     & \textbf{0.308}\cellcolor{iccvblue!15} & 0.724\cellcolor{iccvblue!15}           & \textbf{4.069}\cellcolor{iccvblue!15} & \textbf{73.316}\cellcolor{iccvblue!15} & 14.5M\cellcolor{iccvblue!15}                                 & 0.60 \cellcolor{iccvblue!15}                                    \\
        \bottomrule
    \end{tabular}
    \caption{Quantitative comparison of different control methods across various
    conditioning tasks.}
    \label{tab:comparison}
\end{table*}

Table~\ref{tab:comparison} presents quantitative results across various conditioning
tasks. OminiControl2 better preserves generation quality compared to the baseline
while significantly reducing computational overhead. For multi-condition tasks,
our method reduces conditional overhead by 96.2\% (from 24.38s to 0.93s) while achieving
competitive FID scores. The naïve caching approach shows efficiency gains but
suffers from quality degradation, particularly in deblur tasks. These results
confirm that OminiControl2 successfully addresses computational bottlenecks while
better preserving generation quality across diverse conditioning tasks.

\paragraph{Comparison with other control methods.}
As shown in Table~\ref{tab:comparison}, OminiControl2 maintains the superior
generation quality of its predecessor while significantly reducing computational
overhead. These results confirm that our method successfully preserves the effectiveness
of the original OminiControl framework while achieving computational efficiency comparable
to traditional control methods like ControlNet~\cite{zhang2023adding}.
Furthermore, our approach inherits the parameter efficiency advantage of OminiControl,
requiring only 14.5M additional parameters, making it both computationally
efficient and parameter efficient for conditional generation tasks.

\subsection{Empirical studies}
\label{sec:exp:empirical}

\begin{figure}[!t]
    \centering
    \includegraphics[width=\linewidth]{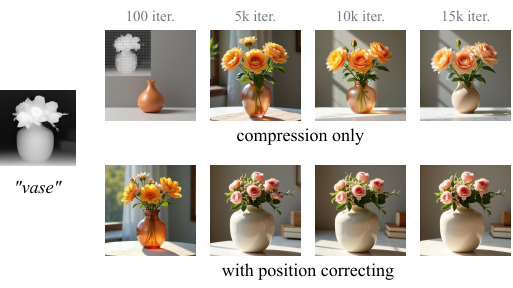}
    \caption{Comparison of image generation results with and without position
    correcting during training.}
    \label{fig:efficiency}
\end{figure}

\paragraph{Position correcting for compressed tokens.}
To evaluate the effectiveness of the proposed position correcting mechanism for compressed
condition image, we compared image generation with and without this technique. Figure~\ref{fig:efficiency}
shows that models trained with position correcting produce significantly better outputs.
Without position correction, the model struggles to establish proper spatial correspondence
between compressed condition tokens and the generated image, resulting in
structural inconsistencies.



\paragraph{Ablation studies.}
Our ablation studies (white rows in Table~\ref{tab:quantitative}) reveal the individual
contributions of our two components. Feature Reuse (F.R.) alone offers substantial
speedup (84.7\% overhead reduction) with minimal quality degradation across most
tasks. Compact Token Representation (C.T.R.) alone provides a balanced tradeoff,
reducing overhead by approximately 65\% while maintaining competitive metrics, notably
achieving better MUSIQ scores for deblurring (72.26 vs. 69.47) than our complete
model. For inpainting, our full method reduces FID from 12.07 to 11.71 while
cutting overhead by 90.1\%. These results demonstrate that both techniques are
complementary, with their combination achieving the optimal balance between quality
and efficiency.

    \section{Conclusion}
    We presented OminiControl2, an efficient framework for controllable image generation
    with Diffusion Transformers. By introducing compact token representation and
    conditional feature reuse, our approach achieves speedups of up to 5.9×
    across various control tasks. These optimizations effectively address the computational
    bottlenecks in multi-conditional generation, enabling practical deployment
    of complex control scenarios. Our work demonstrates that computational efficiency
    can be improved simultaneously by eliminating redundancies in the diffusion process,
    making powerful controllable generation more accessible for real-world
    applications.

        \section{Acknowledgment}
We would like to acknowledge that the computational work involved in this research work is partially supported by NUS IT’s Research Computing group using grant numbers NUSREC-HPC-00001.

    { \small \bibliographystyle{ieeenat_fullname} \bibliography{main} }
\end{document}